\documentclass{article}

\usepackage{calligra}
\usepackage{algorithm}
\usepackage{algorithmicx}
\usepackage{algpseudocode} 
\usepackage{amsmath} 
\usepackage{algorithm}
\usepackage{algpseudocode}

\errorcontextlines\maxdimen

\makeatletter
\newcommand*{\algrule}[1][\algorithmicindent]{\makebox[#1][l]{\hspace*{.5em}\thealgruleextra\vrule height \thealgruleheight depth \thealgruledepth}}%
\newcommand*{\thealgruleextra}{}
\newcommand*{\thealgruleheight}{.75\baselineskip}
\newcommand*{\thealgruledepth}{.25\baselineskip}

\newcount\ALG@printindent@tempcnta
\def\ALG@printindent{%
	\ifnum \theALG@nested>0
	\ifx\ALG@text\ALG@x@notext
	\else
	\unskip
	\addvspace{-1pt}
	\ALG@printindent@tempcnta=1
	\loop
	\algrule[\csname ALG@ind@\the\ALG@printindent@tempcnta\endcsname]%
	\advance \ALG@printindent@tempcnta 1
	\ifnum \ALG@printindent@tempcnta<\numexpr\theALG@nested+1\relax
	\repeat
	\fi
	\fi
}%
\usepackage{etoolbox}
\patchcmd{\ALG@doentity}{\noindent\hskip\ALG@tlm}{\ALG@printindent}{}{\errmessage{failed to patch}}
\makeatother

\newbox\statebox
\newcommand{\myState}[1]{%
	\setbox\statebox=\vbox{#1}%
	\edef\thealgruleheight{\dimexpr \the\ht\statebox+1pt\relax}%
	\edef\thealgruledepth{\dimexpr \the\dp\statebox+1pt\relax}%
	\ifdim\thealgruleheight<.75\baselineskip
	\def\thealgruleheight{\dimexpr .75\baselineskip+1pt\relax}%
	\fi
	\ifdim\thealgruledepth<.25\baselineskip
	\def\thealgruledepth{\dimexpr .25\baselineskip+1pt\relax}%
	\fi
	\State #1%
	\def\thealgruleheight{\dimexpr .75\baselineskip+1pt\relax}%
	\def\thealgruledepth{\dimexpr .25\baselineskip+1pt\relax}%
}

\usepackage[preprint, nonatbib]{nips_2018}

\usepackage{float}

\usepackage[square,numbers]{natbib}
\usepackage[utf8]{inputenc} 
\usepackage[T1]{fontenc}    
\usepackage{hyperref}       
\usepackage{url}            
\usepackage{booktabs}       
\usepackage{amsfonts}       
\usepackage{nicefrac}       
\usepackage{microtype}      
\usepackage{graphicx}
\usepackage{tabularx}
\usepackage{hyperref}

\title{Few-shot Multi-hop Question Answering over Knowledge Base}

\author{
  Meihao Fan \\
  \texttt{631907060104@mails.cqjtu.edu.cn} \\
\And
 Lei Zhang* \\
  \texttt{zhangleicqjtu@163.com} \\
\And
 Siyao Xiao\\
  \texttt{631862020224@mails.cqjtu.edu.cn} \\
  \And
 Yuru Liang \\
    \texttt{2019051507131@stu.cqnu.edu.cn} \\
}

\begin{document}
\maketitle

\begin{abstract}
KBQA is a task that requires to answer questions by using semantic structured information in knowledge base. Previous work in this area has been restricted due to the lack of large semantic parsing dataset and the exponential growth of searching space with the increasing hops of relation paths. In this paper, we propose an efficient pipeline method equipped with a pre-trained language model. By adopting Beam Search algorithm, the searching space will not be restricted in subgraph of 3 hops. Besides, we propose a data generation strategy, which enables our model to generalize well from few training samples. We evaluate our model on an open-domain complex Chinese Question Answering task CCKS2019 and achieve F1-score of 62.55\% on the test dataset. In addition, in order to test the few-shot learning capability of our model, we ramdomly select 10\% of the primary data to train our model, the result shows that our model can still achieves F1-score of 58.54\%, which verifies the capability of our model to process KBQA task and the advantage in few-shot Learning.
\end{abstract}

\section{Introduction}

Due to the proliferation of Artificial Intelligence(AI), smart systems\cite{ref1} have made significant achievements in communication\cite{ref2}\cite{ref3}\cite{ref4}\cite{ref5}\cite{ref6} and information extraction\cite{ref7}\cite{ref8}. Since a sophisticated smart system can bring much convenience and efficiency , the research in this field has attracted extensive attention from academic and industrial circles.

\begin{figure}[!htb]
		\centering
		 \includegraphics[width=0.75\linewidth]{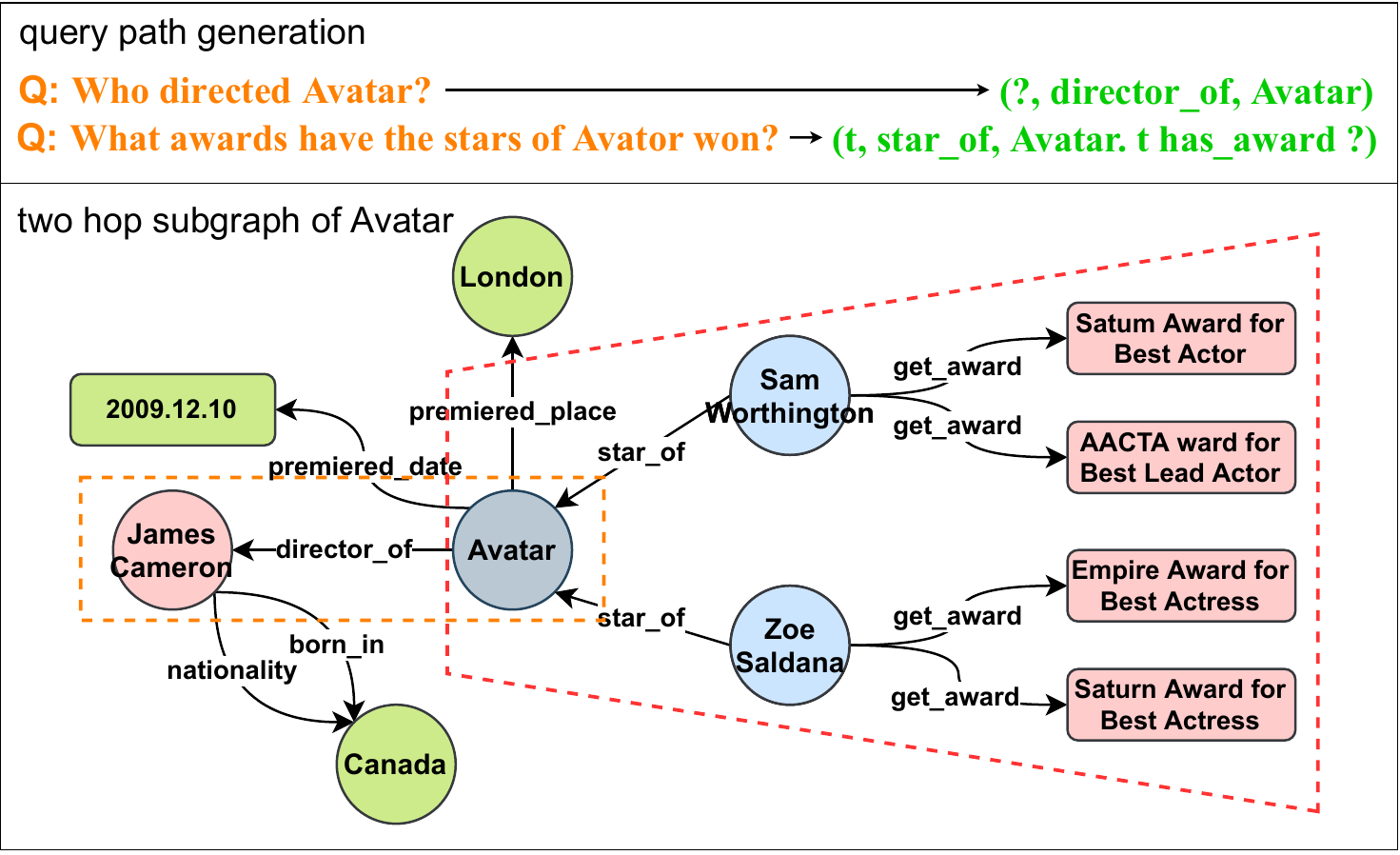}
		\caption{The subgraph of KB and two query paths corresponding to questions of different types}
		\label{fig:two-question}
		\end{figure}

A KBQA system aims to Answer Questions(QA) by understanding the semantic structure and extract the answers in large Knowledge Bases(KB).Recently, tremendous KBQA models are proposed to effectively utilize KB to answer `simple' questions. Here `simple' refers to questions that can be answered with a single predicate or a predicate sequence in the KB. For instance, ``Who directed Avatar?'' is a simple question due to its answer can be obtained by a single triplet fact query (?, director\_of, Avatar). To answer such questions, plenty of Rule-based\cite{ref9}, Keyword-based\cite{ref10} and Synonym-based methods\cite{ref11}\cite{ref13}\cite{ref13}\cite{ref14} have been proposed. However, questions in real life are usually more complex which can only be answered correctly by multi-hop query path with constraints. As is shown in Figure\ref{fig:two-question}, for answering a complex question, a sequence of operations need to be generated, including multi-hop query and answers combination. Recently using KB to Answer such Complex Questions(KBCQA) has attracted growing interests prodigiously\cite{ref15}. Previous state-of-art KBCQA models can be categorized into a taxonomy that contains two main branches, namely Information Retrived-based(IR-based) and Neural Semantic Parsing-based(SP-based). The IR-based model\cite{ref16}\cite{ref17}\cite{ref18}\cite{ref19}\cite{ref20} first recognizes topic entities in the natural language and links them to Node Entities in Knowledge Base. Then all nodes surrounding around the topic nodes are regarded as candidate answers, and a score function is used to model their semantic relevance and predict the final answers. Methods based on Semantic Parsing\cite{ref21}\cite{ref22}\cite{ref23}\cite{ref7}\cite{ref24}) usually includes a Seq2Seq module which converts natural languages into executable query languages and a Executor Module which executes the generated logical sequence on KB to obtain the final answers. 

	However, although the state-of-art models have made great achievements, several challenges still exist. Firstly, the dependency of annotated data is a thorny problem for SP-based models, which is usually settled by using a breadth-first search(BFS) to produce pseudo-gold action sequences and adopting a Reinforce-Learning(RL) algorithm(\cite{ref25, ref26}). Yet since BFS will inevitably ignore many other plausible annotations and RL usually suffers from several challenges, such as sparse reward and data inefficiency, the research of SP-based models are immensely hindered. Secondly, both IR-based and SP-based methods suffer from the Large Searching Space. For better performance on KBCQA task, large KBs, such as Wikidata\cite{ref27} or FreeBase\cite{ref28} are usually needed. Although these KBs contain comprehensive knowledge, they also bring vast search space when searching a query path with more than 3 hops. We record the average number of relations in one hop and multi-hop subgraphs of a topic entity in our training dataset. It is shown that in one-hop subgraphs, the average number is 514.908 while in 2-hop and 3-hop subgraphs, it grows to 1919.446 and 6408.070 respectively. This exponential growth of generated candidate tuples makes it expensive and difficult for caculation. Thirdly, most previous work requires Large KBCQA datasets to train their model, such as ComplexWebQuestions\cite{ref29} and QALD\cite{ref30}. However, this Large Datasets are usually in English, hindering research in more realistic settings and in languages other than English.
	
	To solve the three problems above, we propose a template-based model consisting of Question Classification, Named Entity Recognition, Query Paths Generating and Path Ranking Module. Our contribution can be categorized into three fields:
	
	1. We propose a data-efficient model equipped with a Pre-Trained Language Model BERT\footnotemark which can achieve high performance but only use tiny amount of data. Thus, our model can be utilized to process KBQA task in some languages without large KBQA datasets.
\footnotetext[1]{For better performance, We select ERNIE\cite{ref31} as our pre-trained language model.}
	
	2. By adopting Beam Search algorithm and using BERT to score for each searching branch, the Spatial Complexity and Time Complexity have been greatly dropped but the generating accuracy still remains competitive.
	
	3. We put forward a method to construct artificial data on pre-defined schemas of query graphs, allowing our model to process questions with novel categories which are excluded by training set.

With the utilize of pre-trained language model BERT and pre-defined schemas of query graphs, our model can effectively extract and filter the query tuples for a complex question. Also, we adopt Beam Search algorithm to relieve the exponential growth with increasing hops, which make it possible to handle multi-hop questions.

This paper is organized as follows: In Section~\ref{sec:related-word}, we review works on NER and Beam Search, which are the basis of our experiments. In Section~\ref{sec:our-method} we present the overall architecture, and then introduce each key component in detail. In Section~\ref{sec:our-method}, we demonstrate the evaluated models and the methodology used to generate the sentence embeddings. In Section~\ref{sec:experiments}, we describe the experimental setup and evaluation of the proposed model. Finally, we summarize the contribution of this work in the Section~\ref{sec:conclusion}.

\section{Related Work \label{sec:related-word}}
Recently, with the rapid development and increasing attention of deep learning, the research on natural language processing has made great process. Especially when supported by emerging word embedding technologies and pretrained language models, the effectiveness of knowledge base question answering has been greatly improved. In this section, we will introduce some previous work related to the sub-modules of our model including Named Entity Recognition(NER) and Beam Search algorithm. Besides, some few-shot KBQA models and a template-based model will also be introduced.

Named Entity Recognition is a key component in NLP systems for question answering, information retrieval, relation extraction, etc. Early NER models are mainly based on unsupervised and bootstrapped system\cite{ref32}\cite{ref33} or Feature-engineering supervised task\cite{ref34}\cite{ref35}. Nowadays, researchers tend to use neural network for NER task. NER is often solved as a sequence labeling problem by using the Conditional Random Field(CRF) which requires a set of pre-defined features. Recently, some effective neural network approaches, especially for Bi-directional Long Short-Term Memory, significantly improve the performance of CRF for NER task. Huang et al.\cite{ref30} using two LSTMs to capture past features and future features in sequence tagging task. Then a CRF layer is used to efficiently grasp the sentence level tag information of the sentence. The BiLSTM-CRF is usually employed as the cornerstone of many subsequent improved NER models. BERT BiLSTM CRF\cite{ref36} uses BERT to embed extract rich semantic features into vectors and sends them to the BiLSTM CRF. This model has achieved state-of-art performance in many NER tasks\cite{ref37}.

Beam Search is a common heuristic algorithm for decoding structured predictors. When generating query paths for complex multi-hop questions, we need to consider longer relation path in order to reach the correct answers. However, the search space grows exponentially with the length of relation paths, bringing expensiveness for calculation and storage. The core idea of beam search is to use a score function to keep Top-K candidate relations instead of considering all relations when extending a relation path. Thus, the definition of score function determines the performance of Bean Search. Chen et al.(2019)\cite{ref38}  proposed to keep only the best matching relation with a path ranking module that considers features extracted from topic entities and semantic information of the generated query paths. Lan et al.(2019)\cite{ref39} also keep only one candidate relation using a traditional Siamese architecture where both the question and the candidate paths are each separately encoded into a single vector before the two vectors are matched. The experiment's results of this two models show little performance dropped but with significant reduction in Spatial Complexity and Time Complexity.

Since the expensiveness of constructing the annotated datasets, several works have been focused on few-shot learning for KBQA task. Chada et al.(2021) \cite{ref40} proposed a simple fine-tuning framework that regards the query path generation as a text-to-text task. By leveraging a pre-trained sequence-to-sequence models, their method outperforms many state-of-art models with an average margin of 34.2 F1 points on various few-shot settings of multiple QA benchmarks. Hua et al. (2020) \cite{ref41} proposed a Semantic-Parsing based method using BFS to find the pseudo-gold annotation of a question and learning a Reinforcement-Learning(RL) policy to generate a query sequence for obtaining the final answer.

Our model is most inspired by a template-based Chinese KBCQA model proposed by Wang et al.\cite{ref42}. They use a pipeline method including a NER module, a query paths generating module and candidate tuples ranking module and process the question step by step. In NER module, they attach the BiLSTM CRF layer with a BERT layer to better understand the semantic information in the question, which gets quite high accuracy in topic entities recognition. Then, they extend one or two relations from the topic entity to generate the query paths and adopt Bridging technology to process questions with multiple entities. Finally, a candidate query paths ranking module is carefully designed to select the final query path. The differences between their work and our model are that we process the one-entity and multi-entity questions separately with a Question Classification module and predefine a set of query schema to restrict the searching space. On the predefined query pattern, we use a strategy to construct artificial questions which improve the ability of the classification model for few-shot learning.  Moreover, we adopt Beam Search algorithm when generating query paths, which helps us achieve comparable performance but only using 10$\%$ resource of calculation and storage.

\begin{figure}[!htb]
		\centering
		 \includegraphics[width=\linewidth]{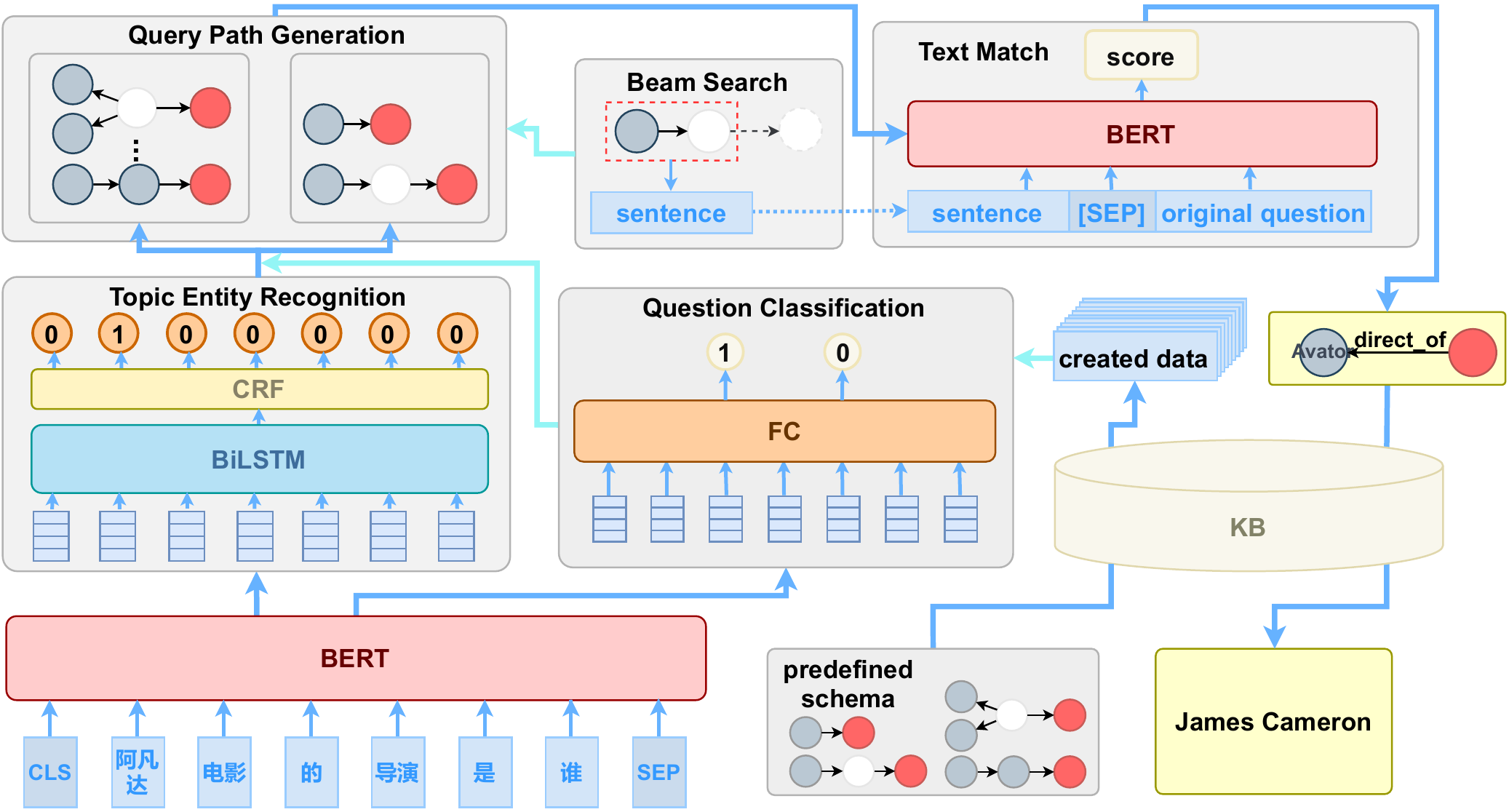}
		\caption{Basic framework of our model.}
		\label{fig:architecture}
		\end{figure}

\section{Our Method}
\label{sec:our-method}
In this section, we will present the overall architecture(shown in Figure\ref{fig:architecture}), and then introduce each key component of the proposed model in detail.

\subsection{Method Overview}

The general idea behind our method is to process the question step by step. Given a question, we first encode it with a BERT layer and then the representations will be passed to an Entity Linking module(Sec.\ref{sub:node-extractor}) of BERT-BiLSTM-CRF layer and a Question Classification Module(Sec.\ref{sub:question-classify}) trained with extra manually constructed samples(Sec.\ref{subsec:data-construction}). With the recognized topic entities and a specific category the question belongs to, we can refer to a more precise schema(Sec.\ref{subsec:predefine-query-schema}) to generate the query path in a narrower searching space. However, since the query graph of a complex question may involve multiple relations, such simple generating program will bring intolerable Time Complexity and Spatial Complexity, and bring calculating burden to the Candidate Tuples Ranking module. To solve this, we adopt a heuristic algorithm for graph search(Sec.\ref{sec:beam-search}) based on a pre-trained text-match model, which greatly decreases the number of candidate query paths. Afterwards, a Candidate Tuples Ranking module is designed to sift out the final path using the above PTM-TextMatch model. By executing the golden query tuple, we can retrieve the answer in Knowledge Base.

Besides, we are not to search aimlessly in KB when generating query subgraph. Instead, we refer to a set of pre-defined schemas of all possible query graphs in complex question answering. This policy will not only narrow the searching space significantly but also provide a semantic framework for reference when constructing artificial questions.

\subsection{Node Extractor \label{sub:node-extractor}}
The main goal of this module is to identify topic entities in the question. This module includes Tokenization with dictionaries, Named Entity Recognition(NER) and Entity Linking.

 \textit{1) Tokenize} : Different from English Tokenize, Chinese Tokenizing usually use dictionaries as a supplementary to tokenize Chinese question text into Chinese words. In this paper, we use a dictionary provided by CCKS consisting of all subjects in KB, all entities and their mentions in Mention Dictinary.

\textit{2) Named Entity Recognition}: In the NER module, we encode the question with BERT layer, and then pass it through a BiLSTM to capture the infomation of context and a CRF layer to predict label of each token.
Let us use $Q=(t_1,t_2,t_3,...,t_n)$ to represent a tokenized question. We put Q into a BERT layer to encode representations with semantic knowledge. Next, the representations $Q={X_i}_{i=1}^{\mid Q \mid}$ are passed through a BiLSTM layer and CRF layer\cite{ref28}. 

For each input token, the context information is captured by two LSTMs, where one capture information from left to right, the other from right to left. At each time step $t$, a hidden vector $\overrightarrow{h_t}$(from left to right) is computed based on the previous hidden state $\overrightarrow{h_{t-1}}$ and the input at the current step $x_t$. Then The forward and backward context representations, generated by $\overrightarrow{h_t}$ and $\overleftarrow{h_t}$ respectively, are concatenated into a long vector which we represent as $h_t=[\overrightarrow{h_t}:\overleftarrow{h_t}]$. The basic LSTM function is defined as follow:
   \begin{equation}
    \left[  \begin{array}{c}\tilde{c}_t \\ f_t \\ o_t \\ i_t\end{array}\right] =\left[\begin{array}{c}\sigma \\ \sigma \\ \sigma \\ \tanh \end{array} \right] \Big ( W^T \left[\begin{array}{c} x_t \\ h_{t-1} \end{array}\right] + b\Big ) 
    \end{equation}
    \begin{equation}
     	c_t=i_t\odot \tilde{c}_t+f_t\odot c_{t-1}
     	\end{equation}
     	  \begin{equation}
       h_t=o_t \odot tanh(c_t)
    \end{equation}
  where $W^T$ and $b$ are trainable parameters; $\sigma()$ is the sigmoid function; $i_t , o_t, f_t$ indicating input, output and forget gates respectively; $\odot$ represents the dot product function; $x_t$ is the input vector of the current time step.
  
  The output vectors of the BiLSTM contain the bidirectional relation information of the words in a question. Then we adopt CRF to predict labels for each word, considering the dependencies of adjacent labels. The CRF is the Markov random field of Y given a random variable X condition and included a undirected graph G where Y are connected by undirected edges indicating dependencies. Formally, given observations variables $H={h_i}_{i=1}^{\mid Q \mid}$, and a set of output values $y\in {0,1}$, where $y=1$ means the corresponding token is a topic entity and $y=0$ is not. CRF defines potential fucntions as below:
  $$p(y\mid h)=\frac{1}{Z_h}\prod_{s\in S(y,h)}\phi_s(y_s, h_s)$$
  where $Z_h$ is a normalization factor overall output values, $S(y,h)$ is the set of cliques of $G$, $\phi_s(y_s, h_s)$ is the clique potential on clique $s$. 
  
  Afterwards, in the BiLSTM-CRF model, a softmax over all possible tag sequences yields a probability for the sequencey. The prediction of the output sequence is computed as follows: 
  $$y_*=argmax_{y\in\{0,1\}}\sigma(H,y)$$
  $$\sigma(H,y)=\sum_{i=0}^n A_{y_i,y_{i+1}} + \sum_{i=0}^n P_{i,y_{i}}$$

  where $A$ is a matrix of transition scores, $A_{y_i,y_{i+1}}$ represents the score of a transition from the tag $y_i$ to $y_{i+1}$. $n$ is the length of a sentence, $P$ is the matrix of scores output by the BiLSTM network, $P_{i,y_{i}}$ is the score of the $y_i^{th}$ tag of the $i^{th}$ word in a sentence.
  
\textit{3) Entity Linking}: In this module, we link the recognized named entity to the entity in KB and select a set of candidate topic entities with a Mention Dictionary. The Mention Dic is a dictionary provided by CCKS Sponsors describing mapping relations from mentions to node entities. After obtaining mentions of entities in a question, we correspond them to relevant node entities. Then we need to extract helpful features from the mentions and entities to select the potential candidate entities. In this work, we extract six features as below: The Length of Entity Mention($f_1$), The TF value of Entity Mention($f_2$), The Distance Between the Entity Mention and Interrogative Word($f_3$), Word Overlap Between Question and Triplet Paths($f_4$), and Popularity of Candidate entities($f_5$). The popularity is calculated as $\sqrt{k}$, where $k$ represents the number of relation path the candidate entity has within 2 hop graph. We assume that a entity with larger $f_1$, $f_2$, $f_4$, $f_5$ and smaller $f_3$, are more likely to be a topic entity.

This six features will be calculated and put into a linear weighing layer to output relative scores. Entities with $Top\ k$ score build the candidate entities set.

      The score is calculated using the following fuction:
$$s=w_1\cdot f_1 + w_2\cdot f_2 + w_3\cdot f_3 + w_4\cdot f_4 + w_5\cdot f_5$$
where $f_i$ represents the $i^{th}$ feature and $w_i$ represents the corresponding weight.

\subsection{Question Classification \label{sub:question-classify}}
In order to improve the efficiency of our model, we use a pre-trained language model BERT to classify the complex questions into two categories: one topic entity question and multi-entity question, and process each of them separately. In one entity question, predicted paths usually extend from the topic entity with one relation or a sequence of relation hops. While in multi-entity questions, correct answers can only be obtained accurately by executing the query paths extended from several topic entities in the question. For instance, the question ``Whose husband is the director of Avator?'' is one-entity question because its query paths (?, wife\_of, t. t, director\_of, Avator) can be extracted from the ``Avator'' through the relations ``director\_of'' and ``wife\_of'' and the transitional entity t. Meanwhile, ``Which actors in Avator born in British?'' is a complex question because the correct query paths can only be generated from the entity ``Avator'' and ``British'' respectively through the relations ``actor\_of'' and ``born\_in''. In addition, we generate artificial questions in a semantic structured form to improve the performance of our classification model. The detailed implementation will be represented in subsection\ref{subsec:data-construction}.

Given a question, we encode it with words encoding, position encoding and segment encoding, and attach a special token $[CLS]$ at the beginning of a question to separate different sentences. Then the semantic information will be captured with a multi-head attention system and a Dense Layer will be attached to obtain the prediction.

\subsection{Predefine the Query Schema\label{subsec:predefine-query-schema}}

The golden key to solving the KBCQA task is to map entities of a question into a specific query graph. A Semantic Parsing-based model transfers the KBQA task into a Seq2Seq task. By feeding the model with numerous annotated data, SP-based model can understand the semantic framework of a question and refine corresponding query graph. An Information Retrived-based model adopts a different method that searches all query graphs surrounding the extracted topic entities and then uses a Candidate Tuple Ranking Module to sift the final query graphs. However, with limited data, it is challenging to learn the query structure of questions, let alone changing it to an executable action sequence. In this work, we relieve this problem by predefining the schema of query graph and adopt Beam Search to pruning the searching space of multi-hop query paths.

Inspired by Aqqu\cite{ref43}, we propose an inverse solution that we first take a deep insight into numerous Chinese multi-hop questions and propose eight searching schemas for complex questions as shown in Figure\ref{fig:predefined-schema}. By predefining the schema of query graph, our model can benefit from three aspects:

\begin{figure}[!htb]
		\centering
		 \includegraphics[width=0.65\linewidth]{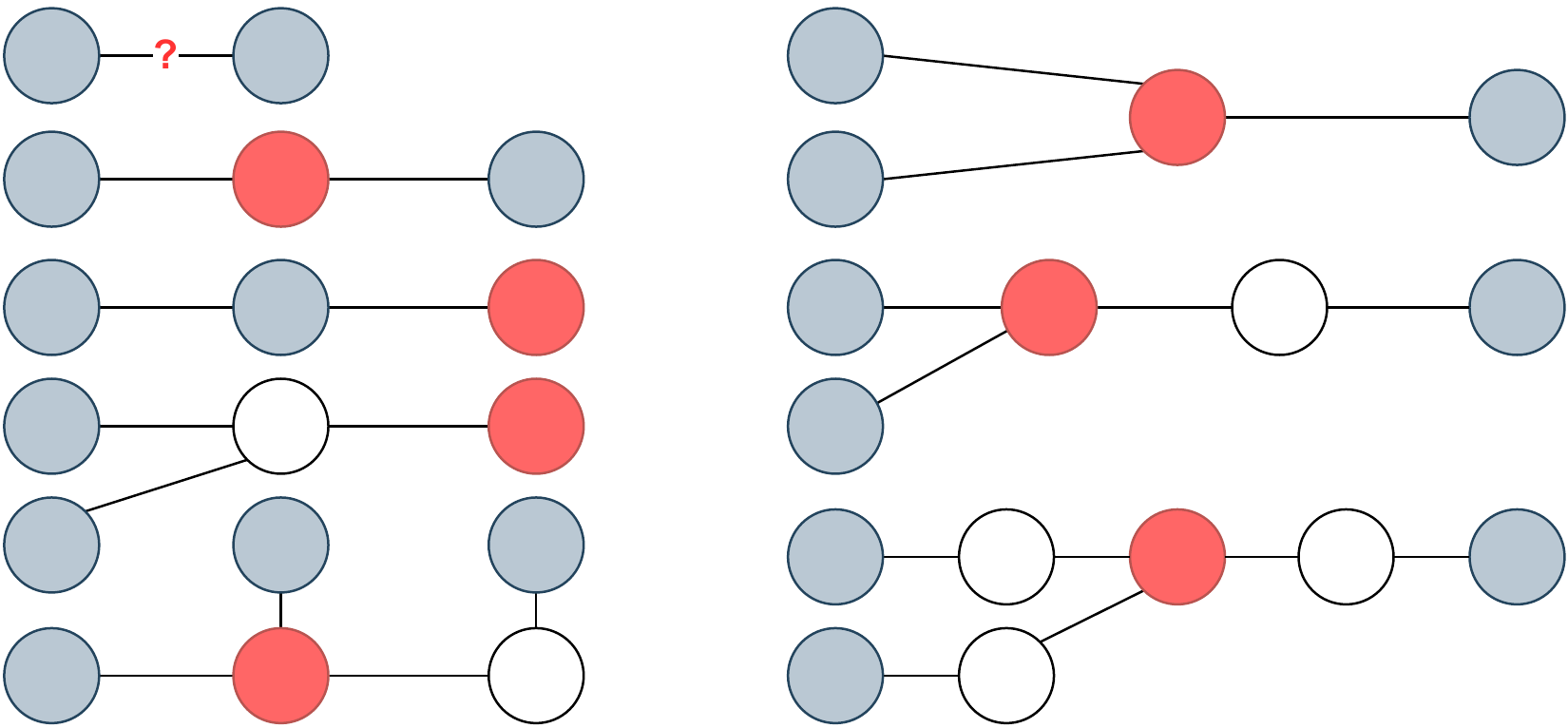}
		\caption{predefined schema of query path. The grey ones represent topic entities we already know. The white one represents transitional entity we need not record and the red one represents the answer we query.
		}
		\label{fig:predefined-schema}
		\end{figure}

a) Predefining the schema introduces prior knowledge, which stipulates the semantic structure of the queried question and greatly prunes the search space.

b) Since the patterns of query tuples are specified, we can easily turn each query tuples into its semantic form and calculate the similarity between the artificial question and real question with a pre-trained language model, which we define as the score of the query path we generate.

c) Extra data can be constructed on the enumerated query schema to train the classification model, which allows the model to learn the basic semantic knowledge of classifying questions.

We assume that the diversity of candidate tuples will lead to poor performance of candidate query path ranking module. Thus, we divide the query schema into two modules according to number of topic entities the query pattern has. When generating query paths, we use two separate modules to generate candidate query paths. For one-entity question, we simply search the sub graph of the topic entity within two relation hops. While for questions of multiple entities, we generate query paths on the searching schemas shown in Figure\ref{fig:predefined-schema}. Let $n$ represent the number of candidate topic entities and $m$ represent the number of true topic entities in a given question. Since combinatorial number $C_n^m$ grows too large when m is greater than 3, we only consider questions containing three or less topic entities.

\subsection{Artificial Data Construction\label{subsec:data-construction}}
For better predicting which class a question belonging to and alleviating the need of labeled training data, we generate substantial artificial questions on the predefined query schemas. In our method, we randomly select a node entity in KB and extend a query path from the entity. When generating a query path, we are not to consider all branches in a random searching schema. Instead, we conduct the algorithm on the predefined query schema which has been introduced in subsection\ref{subsec:predefine-query-schema}. For instance, as for the above question ``Whose husband is the director of Avator?'', the corresponding query schema is ($x$, $r_1$, $t$. $t$, $r_2$, $e$) where $x$ represents the answer, $r_1$ and $r_2$ represent any relations in two hops query path extended from the topic entity $e$ through an intermediate entity $t$. We generate the artificial question by replacing mentions of topic entities(in this example is ``Avator'') and relations(``wife\_of'', ``director\_of'') with mentions of randomly selected node entities and correlated relations. In addition, if the query schema is excluded in training samples, we only need to manually construct a fake question corresponding to the query schema and then execute the above steps.

Since our predefined query schema contains semantic structure for both one-entity and multi-entity questions, our constructed samples can lead the pre-trained language model to converge in a direction which is more compatible with our specific classification task. Besides, the ratio of questions of different query patterns should be carefully controlled in order to improve the generalization of created data. 

Although our constructed questions have some differences from the real questions in semantic expression, our model can still learn extra semantic structure of questions in two classes. In our experiment, we constructed 5k artificial questions and use them to train our classification model. With the help of pre-trained language model, our model can handle some questions that have never shown in training set. As the results in Sec.\ref{sec:experiments} shown, given only 10\% of training data, our model can achieve good performance in classifying the questions.

\subsection{Beam Search \label{sec:beam-search}}

 It is worth to notice that when extending multi-hop relations of the two type questions above, Query Path Generation module often suffers from the vast searching space. To solve this, we adopt a Heuristic algorithm Beam Search algorithm equipped with a pretrained language model BERT to score for each breach of relations, thus we avoid exhaustive search on irrelevant relations. When extending a new relation path at n-step, we try to add the relation $r_n$ to the previous generated query path $R_{n-1}$ and use the strategy introduced in Artificial Data Construction (Sec.\ref{subsec:data-construction}) to transfer the graph into a semantic form $S_n$. Then $S_n$ and original question $Q$ are tokenized and concatenated with a special token $[SEP]$ as below:
 $$input=[CLS]S_n[SEP]Q$$
 This two sentences are fed into a pretrained language model of downstream task to calculate the semantic similarity which represents the score for $r_n$ given a sub query path $R_{n-1}$. The formulation is defined as:
 $$Sco(r_n|R_{n-1})=BERTLayer(input)$$
  At each extending step, we only consider relations with $Top_k$ score for further search, which significantly excluded some irrelavant query branches. The result in Sec.\ref{subsec:exp-beam-search} shows that, by adopting the Beam Search algorithm, the accuracy of query paths generating remains competitive but the number of candidate paths decrease above 80\%. The detailed description is seen in Algorithm\ref{algorithm1}.

\begin{algorithm}[ht!]  
	\renewcommand{\algorithmicrequire}{\textbf{Input:}}
	\renewcommand{\algorithmicensure}{\textbf{Output:}}
	\caption{Multihop relation extraction. For each query schema, we generate a set of candidate query paths $P^{(T)}$, where T represents the hop number of the schema.}  
	\label{algorithm1}
	\begin{algorithmic}[1] 
		\Require $KB$, question $q$, topic entity set $E$, number of hops $T$
		\Ensure  $P^{(T)}$ \\
		\textbf{Initialize:} $P^{(0)}\leftarrow \{e_0\in E\}$
		\For{$t=1,2,..., T$} 
			\myState{$\tilde{P}^{(t)}\leftarrow \phi$} 
			\myState{$\tilde{S}^{(t)}\leftarrow \phi$} 
			\For{$each p\in P^{(t-1)}$}
				\myState{$e_{t-1}\leftarrow tail(p)$}
				\For{$each (e_{t-1}, r, e_t) \in KB$}
					\If{$e_t\in E$}
						\myState{$p'\leftarrow p\oplus(r,e_t)$}
					\Else 
						\myState{$p'\leftarrow p\oplus(r)$}
					\EndIf
					\myState{$\tilde{P}^{(t)}\leftarrow \tilde{P}^{(t)}\cup \{p'\}$}
					\myState{$\tilde{S}^{(t)}\leftarrow \tilde{S}^{(t)}\cup \{Sentence(p')\}$}
				\EndFor
			\EndFor
			\myState{score all elements in $\tilde{S}^{(t)}$ and rank all corresponding elements in $\tilde{P}^{(t)}$}
		\EndFor 	
	\end{algorithmic}
\end{algorithm}

\section{Experiments}
\label{sec:experiments}
In this section, we analyse the performance our model achieves on complex question answering with limited training data. We take an insight into each module and conduct ablation experiments to better understand our model.

\subsection{KB and Datasets}
Our model uses a open-domain KB PKU-Base, which adopts Resource Description Framework(RDF) as their data format and contains billions of SPO (subject, predicate, object) triples.\cite{ref30}, as is shown in table\ref{number-of-data-in-KB}. We train and evaluate our model on CCKS datasets, which contain 2298, 766, 766 pairs of questions.

\begin{table}[h]
\centering
\caption{number of triples, entity type and entity linking in PKU-Base}
\label{number-of-data-in-KB}
\scalebox{0.9}{
\begin{tabular}{lccc}
\toprule
\textbf{type}    &\textbf{triples}    & \textbf{entity type}  & \textbf{entity linking}  \\
\midrule
\textit{number of data}       & 61,006,527 &    25,182,627   & 13,930,117 \\

\bottomrule
\end{tabular}}
\end{table}

\subsection{Entity Linking}
In Entity Linking module, we remove each feature of candidate entities to observe the influence on the performance of Entity Linking models. The left column is disassembled model and the right is its recall of recognizing topic entities.

\begin{table}[!htb]
\caption{Results of ablation experiments in Entity Linking Module.}
  \label{entity-linking}
  \centering
    \scalebox{1}{
  \begin{tabular}{l|cc}
\toprule
\textbf{type}    & one-entity & multi-entity  \\
\midrule
\textit{Baseline}      & 0.848   & 0.726  \\
\midrule
w/o $f_1$ & 0.841    & 0.733  \\
w/o $f_2$ & 0.848    & 0.721  \\
w/o $f_3$ & 0.843    & \textbf{0.744}  \\
w/o $f_4$ & 0.838    & 0.706  \\
w/o $f_5$ & \textbf{0.849}    & 0.637   \\
    \bottomrule
  \end{tabular}
  }
  
\end{table}

As is shown in table\ref{entity-linking}, without $f_3$, the recall of multi-entity questions surprisingly increased while accompanied with a sacrifice of accuracy for one-entity questions. Samely, without $f_5$, the topic entity extracting accuracy for questions of one topic entity increases, but the accuracy for multi-entity question drops. Moreover, excluding any of other features, the performance of Entity Linking model drops, which verifies their contribution for this module. Based on the results, we can modify the Entity Linking module by discarding feature $f_5$ in one-entity questionss entity linking stage while only considering $f_1$, $f_2$, $f_4$, and $f_5$ when processing multi-entity questions. This will be included in our further study.

\subsection{Question Classification}
In this module, we construct 5k artificial data based on the predefined query graphs, and attached them to the training datasets. In order to evaluate the learning capability of our model on small amount of data, we train our model on 10\%, 50\% and 100\% randomly selected samples of primary training datasets, and compare their performance with those additionally attached with certain number of created training samples.

Notably, When adding the constructed samples, we should carefully control the quantity according to the number of primary training samples. For one thing, negligible improvement of the learning ability can be brought, if the quantities of the added samples are too small. For the other, Adding too many constructed data will bring Knowledge Noise, which lead the model to learn a distribution far away from the primary datasets. In our experiment, for 10\%, 50\% and 100\% primary training data, we add 0.05k, 0.5k and 3.75k manually constructed samples respectively. The result is illustrated in table\ref{question-classification}.

\begin{table}[!htb]
  \caption{We evaluate our model on primary training datasets, where created samples are excluded.}
  \label{question-classification}
  \centering
    \resizebox{0.5\textwidth}{!}{
  \begin{tabular}{l|ccc}
\toprule
\textbf{data}    & train & valid & test  \\
\midrule
\textit{10\%+raw data}      & 82.90   & 84.31 & 80.13  \\
\textit{10\%+created data}      & \textbf{87.51}   & \textbf{89.54} & \textbf{82.75}  \\
\midrule
\textit{50\%+raw data}      & 94.95   & 93.99 & 88.50  \\
\textit{50\%+created data}      & \textbf{93.86}   & 83.86 & \textbf{89.41}  \\
\midrule
\textit{100\%+raw data}      & 97.39   & 95.42 & 88.76  \\
\textit{100\%+created data}      & \textbf{99.09}   & \textbf{95.45} & \textbf{91.11}  \\
    \bottomrule
  \end{tabular}
  }
\end{table}

From the above table, we find that when attached with manually constructed samples, our model’s performance has improved on both partial and whole primary data. Our strategy can bring more significant improvement especially when given a small amount of training data. Moreover, we can see an obvious improvement of the prediction on  training datasets, which indicates appropriate number of created samples can make the model better fit the distribution of training data.

We owe the model's out performance to the introduction of prior knowledge. Due to the diversity of the samples in datasets, the test set may contain questions whose semantic structures have not appeared in training set. In this zero-shot or few-shot situation, the model may have difficulty predicting the correct class. However, with additional created samples, our model can learn the predefined semantic structures. If these structures appear in test sets while not included by training set, the performance of our model will be improved. Thus, our model may need more steps to converge.

\begin{figure}[!htb]
		\centering
		 \includegraphics[width=0.8\linewidth]{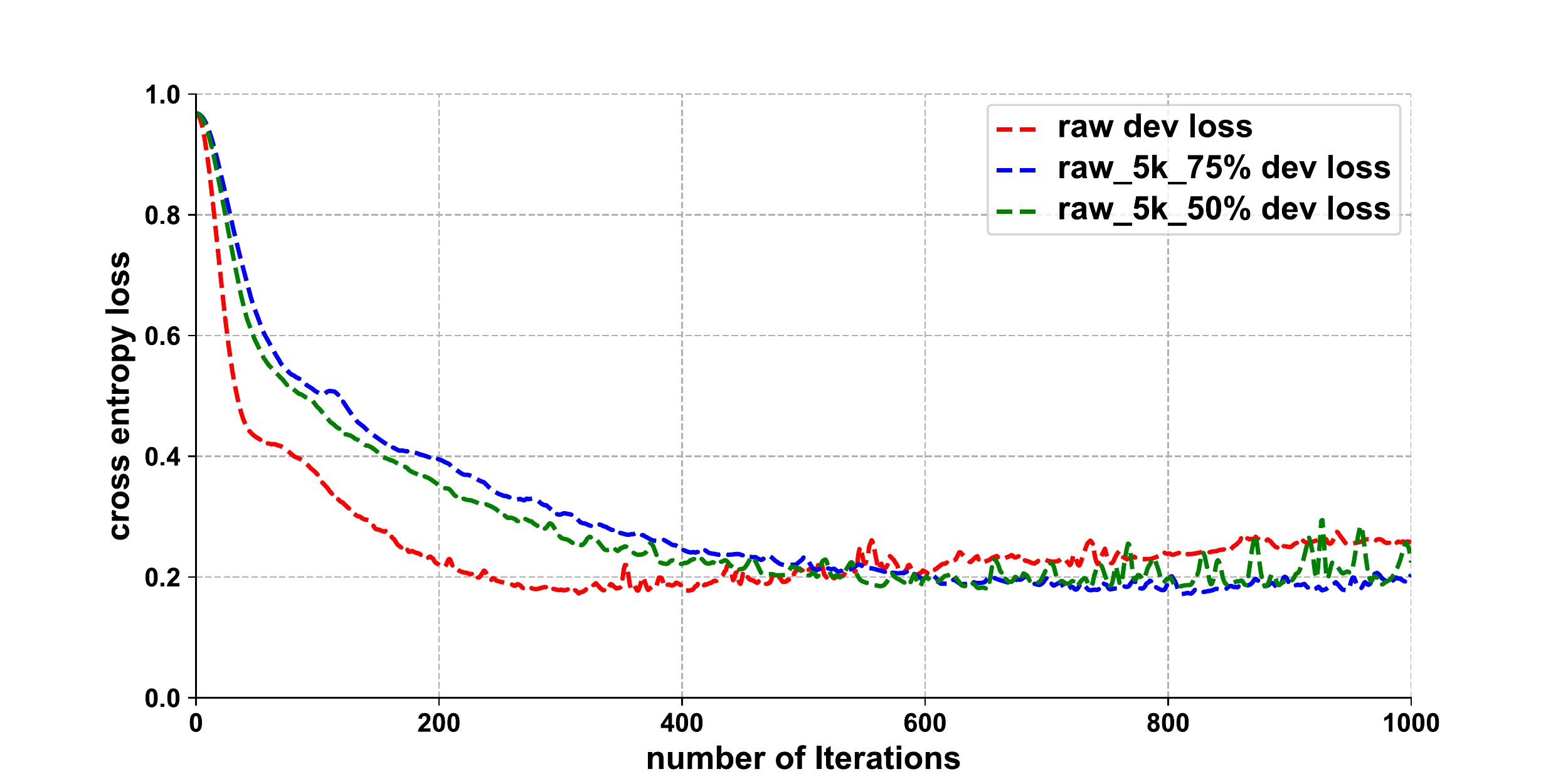}
		\caption{loss of classification model when training with 50\%, 75\% and 100\% created data.}
		\label{fig:100acc}
		\end{figure}

To verify the idea, we record the loss of each iteration when training with total primary data attached with 0\%, 50\% and 75\% created data, shown as figure \ref{fig:100acc}.

We find that when training primary data attached with 0\%, 50\% and 75\% created data, our model converges at about 280, 550 and 760 steps respectively, which indicates that with more created data, the model needs more iterations to converge.

\subsubsection{Beam Search\label{subsec:exp-beam-search}}
For better exhibiting the effect of Beam Search(BS), we select 653 questions whose query path containing 2 hops of relations to test our methods. In the experiment, we design the benchmark by enumerating all the query paths within two-hop relations of the topic entity and recording the average number of query paths N. Notably, we only use BS algorithm at first hop, while searching for the second hop, we only extend from the reserved Top-K sub query path filtered by the BS algorithm and keep all the two hops query paths. By setting different beam size, we can observe the influence on the recall and number of generated query paths.

\begin{figure}[!htb]
		\centering
		 \includegraphics[width=0.8\linewidth]{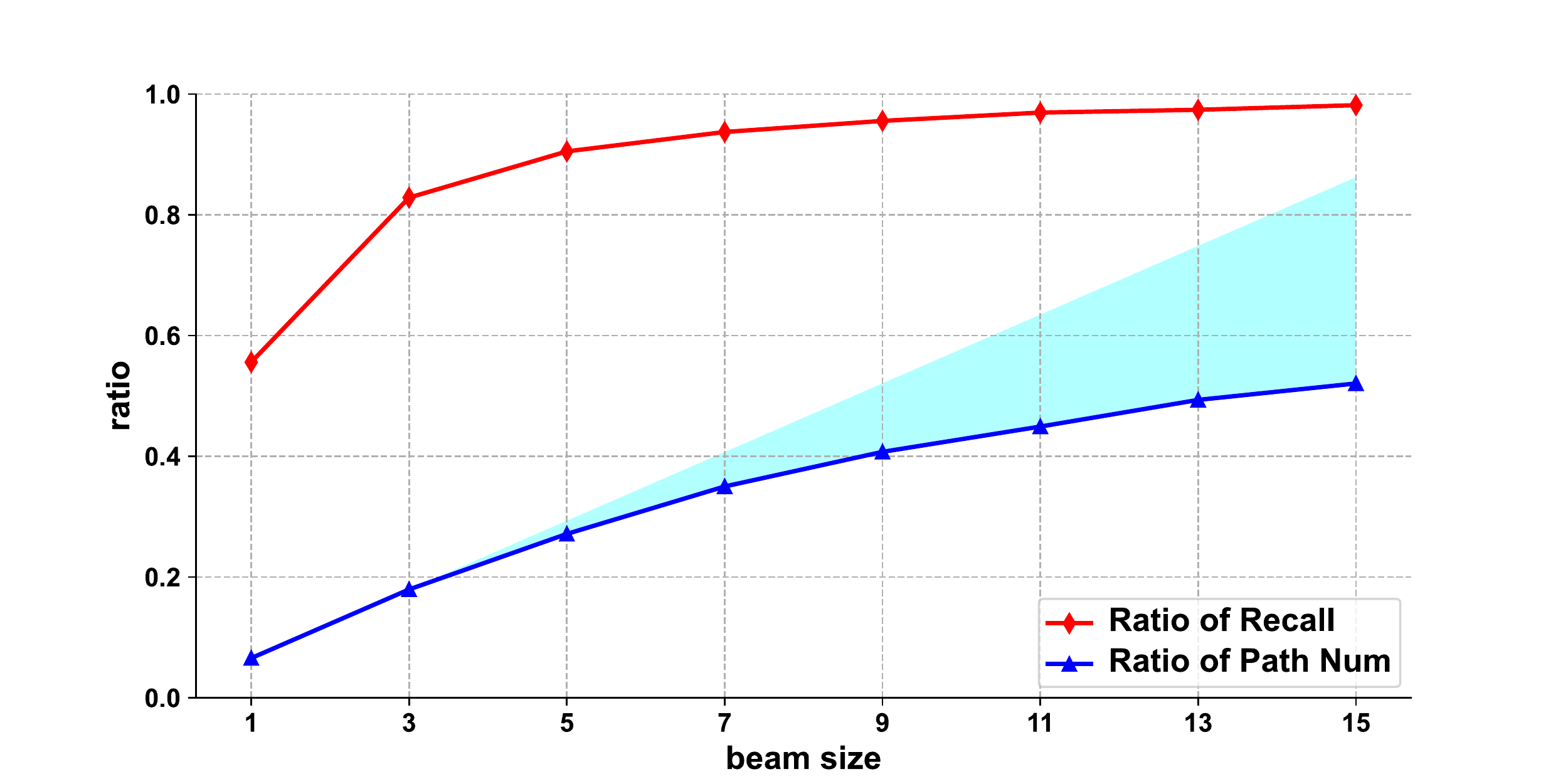}
		\caption{ratio of recall and path numbers when using different beam size.}
		\label{fig:beam-search}
		\end{figure}

Figure\ref{fig:beam-search} shows that a larger beam size will bring an increase in both recall and number of candidate query paths. Through further observation, we notice that the growth of the both indexes slow down. The retarded growth of recall is intelligible. Due to the existence of upper bound, if the beam size is large enough, the recall will infinitely approach to and finally reach 1.0. However, the retarded increasing speed of the number of candidate tuples can illustrate something. When designing the score function for BS, we use a PTLM model to calculate the similarity of generated query paths and primary questions. Thus, the remaining one-hop relations are usually more relevant to the semantic information in the primary question. As the figure shows, extended from a relation with lower semantic score, the second hop tends to generate fewer query paths. Since the relation whose tail has more triples may have more probability to be the component of golden query path, we assume that the language model can be interpreted as a probabilistic model not only in the dimension of words but also in the dimension of query paths.

\subsubsection{Final Result}
We evaluate our model in the CCKS2019 datasets and compare our performance with a start-of-art model proposed by Wang et al\cite{ref42}. Their model first generated all query paths within 2 hops, and adopted Bridging technologies to handle questions with multiple topic entities. In candidate tuple ranking module, Lan\cite{ref5} uses a PTLM model to calculate scores for generated query paths. Notably, their model introduces negative samples when training the Semantic Match model. Besides, since introducing Bridging technology may harm the predicting performance of one-entity questions, they adopt a Literal Match technology to rerank the generated query path. 

We implement their model and run it on a RTX 2080. It must to be pointed out that due to the difference of experimental equipments and subtle distinction of our datasets, the performance we obtain has some discrepancy with Wang proposed. However, since both our model and theirs are trained in the same experiment environment, the comparison is still persuasive.

\begin{table}[!htb]
  \caption{Comparative results between our best model with other models.}
  \label{final-result}
  \centering
    \resizebox{0.6\textwidth}{!}{
  \begin{tabular}{l|cc}
\toprule
Method    & negative sample & Avg $F_1$  \\
\midrule
Wang (baseline)      & 3  & 56.70\%  \\
Wang (Bridging)      & 3  & 58.60\%  \\
Wang (Bridging+literal match)      & 3  & 61.50\%  \\
Wang (Bridging+literal match)      & 1  & 61.10\%  \\
Wang (Bridging+literal match)      & 5  & 59.40\%  \\
\midrule
our model (with 10\% data)      &   & 58.54\%  \\
our model (with 100\% data)			     &   & \textbf{62.55\%}  \\
    \bottomrule
  \end{tabular}
  }
\end{table}

The result shows that our method is data-efficient and high-performed. Only using 10\% data, our model can achieve competitive result. Moreover, when using 100\% data, our model outperforms at over 1.0 point.

\section{Conclusion\label{sec:conclusion}}
This paper proposes a KBQA system equipped with pre-trained language model to handle multi-hop questions. We have shown that our model has the capability of answering multi-hop questions given small amount of data. Besides, experiments have been conducted to demonstrate that, by adopting Beam Search algorithm, we can achieve competitive results with much smaller cost of calculation and storage, which shows the superiority of our model for few-shot KBCQA task.

\section*{Acknowledgements\label{sec:acknowledgements}}
This work was partially supported by Group Building Scientific Innovation Project for universities in Chongqing (CXQT21021). We are also grateful to Ren Li for providing us Experimental equipment.

\section*{Data Availability}
For the access of data mentioned above, please email the corresponding author.

\section*{Conflicts of Interest}
No competing interests exist within this work.

\appendix


\begin{thebibliography}{99}  
\bibitem{ref1}
Sun Z, Wang Y, Cai Z et al. ``A two‐stage privacy protection mechanism based on blockchain in mobile crowdsourcing,'' International Journal of Intelligent Systems, pp. 2058-2080, 2021.

\bibitem{ref2}Wang Y, Cai Z, Zhan Z H et al., ``Walrasian equilibrium-based multiobjective optimization for task allocation in mobile crowdsourcing,'' IEEE Transactions on Computational Social Systems., pp. 1033-1046, 2020.

\bibitem{ref3}
Cai Z P, and He Z B. ``Trading Private Range Counting over Big IoT Data,'' 2019 IEEE 39th International Conference on Distributed Computing Systems(ICDCS), pp. 144-153, 2019.

\bibitem{ref4}
Wang Y, Gao Y, Li Y et al. ``A worker-selection incentive mechanism for optimizing platform-centric mobile crowdsourcing systems,'' Computer Networks, 171: 107144. 2020. 

\bibitem{ref5}
Lan Y, and Jiang J. ``Query Graph Generation for Answering Multi-hop Complex Questions from Knowledge Bases,'' in Proceedings of the 58th Annual Meeting of the Association for Computational Linguistics, pp. 969–974, 2020.

\bibitem{ref6}
Lu Z, Wang Y, Li Y et al. ``Data-Driven Many-Objective Crowd User Selection for Mobile Crowdsourcing in Industrial IoT,'' IEEE Transactions on Industrial Informatics, 2021.

\bibitem{ref7}Yuncheng Hua, Yuan-Fang Li, Guilin Qi et al., ``Less is More: Data-Efficient Complex Question Answering over Knowledge Bases,'' 2020, https://arxiv.org/abs/2010.15881.

\bibitem{ref8}Sharath J S, and Banafsheh R, ``Question answering over knowledge base using language model embeddings,'' 2020 International Joint Conference on Neural Networks (IJCNN)., pp. 1-8, 2020.

\bibitem{ref9}Ou S, Orasan C, Mekhaldi D et al., ``Automatic Question Pattern Generation for Ontology-based Question Answering,'' in Flairs Conference, pp. 183-188, 2008.

\bibitem{ref10}Unger C, Bühmann L, Lehmann J et al., ``Template-based question answering over RDF data,'' in Proceedings of the 21st international conference on World Wide Web, pp. 639-648, 2012.

\bibitem{ref11}Unger C, and Cimiano P. Pythia, ``Pythia: Compositional meaning construction for ontology-based question answering on the semantic web,'' in International conference on application of natural language to information systems, pp. 153-160, 2011.

\bibitem{ref12}M. Yahya, K. Berberich, S. Elbassuoni et al., ``Natural language questions for the web of data,'' in Proceedings of the 2012 Joint Conference on Empirical Methods in Natural Language Processing and Computational Natural Language Learning, pp. 379-390, 2012.

\bibitem{ref13}Unger C, and Cimiano P. Pythia, ``Natural language question answering over RDF: a graph data driven approach,'' in Proceedings of the 2014 ACM SIGMOD international conference on Management of data, pp. 313-324, 2014.

\bibitem{ref14}UZheng W, Zou L, Lian X et al., ``How to
build templates for rdf question/answering: An uncertain graph
similarity join approach,'' in Proceedings of the 2015 ACM SIGMOD international conference on management of data, pp. 1809-1824, 2015.

\bibitem{ref15}Fu B, Qiu Y, Tang C et al., ``A survey on complex question answering over knowledge base: Recent advances and challenges,'' 2020, https://arxiv.org/abs/2007.13069.

\bibitem{ref16} Dong L, Wei F, Zhou M et al., ``Question answering over freebase with multi-column convolutional neural networks,'' in Proceedings of the 53rd Annual Meeting of the Association for Computational Linguistics and the 7th International Joint Conference on Natural Language Processing (Volume 1: Long Papers), pp. 260-269, 2015.

\bibitem{ref17}Hao Y, Zhang Y, Liu K et al., ``An end-to-end model for question answering over knowledge base with cross-attention combining global knowledge,'' in Proceedings of the 55th Annual Meeting of the Association for Computational Linguistics (Volume 1: Long Papers), pp. 221-231, 2017.

\bibitem{ref18}Sun H, Dhingra B, Zaheer M et al., ``Open Domain Question Answering Using Early Fusion of Knowledge Bases and Text,'' in Proceedings of the 2018 Conference on Empirical Methods in Natural Language Processing, pp. 4231-4242, 2018.

\bibitem{ref19}Sun H, Bedrax-Weiss T, and Cohen W W, ``Pullnet: Open domain question answering with iterative retrieval on knowledge bases and text,'' 2019, https://arxiv.org/abs/1904.09537.

\bibitem{ref20}Zi-Yuan Chen, Chih-Hung Chang, YiPei Chen et al., ``UHop:An unrestricted-hop relation extraction framework for knowledge-based question answering,'' 2019, https://arxiv.org/abs/1904.01246.

\bibitem{ref21}Luo K, Lin F, Luo X et al., ``Knowledge base question answering via encoding of complex query graphs,'' in Proceedings of the 2018 Conference on Empirical Methods in Natural Language Processing, pp. 2185-2194, 2018.

\bibitem{ref22}Gaurav Maheshwari, Priyansh Trivedi, Denis Lukovnikov et al., ``Learning to rank query graphs for complex question answering over knowledge graphs,'' in International semantic web conference, pp. 487-504, 2019

\bibitem{ref23}Zhu S, Cheng X, Su S et al., ``Knowledge-based question answering by tree-to-sequence learning,'' Neurocomputing, vol. 372, pp. 64-72, 2020.

\bibitem{ref24}Sun Y, Zhang L, Cheng G et al., ``SPARQA: skeleton-based semantic parsing for complex questions over knowledge bases,'' in Proceedings of the AAAI Conference on Artificial Intelligence, pp. 8952-8959, 2020.

\bibitem{ref25}C. Liang, J. Berant, Q. Le et al., ``Neural symbolic machines: Learning semantic parsers on freebase with weak supervision,'' 2016, https://arxiv.org/abs/1611.00020.


\bibitem{ref26}Ansari G A, Saha A, Kumar V et al., ``Neural Program Induction for KBQA Without Gold Programs or Query Annotations,'' in IJCAI, pp. 4890-4896, 2019.


\bibitem{ref27}D. Vrandeˇci´c, and M. Kr¨otzsch, ``Wikidata: a free collaborative knowledgebase,'' in Communications of the ACM, vol. 27, pp 78-85, 2014.

\bibitem{ref28}K. Bollacker, C. Evans, P. Paritosh et al., ``Freebase: a
collaboratively created graph database for structuring human knowledge,'' in Proceedings of the 2008 ACM SIGMOD international conference on
Management of data, pp. 1247–1250, 2008.

\bibitem{ref29}Talmor A, and Berant J, ``The Web as a Knowledge-Base for Answering Complex Ques-
tions,'' 2018, https://arxiv.org/abs/1803.06643.


\bibitem{ref30}Unger C, Forascu C, Lopez V et al., ``Question answering over linked data (QALD-4),'' Working Notes for CLEF 2014 Conference., 2014.

\bibitem{ref31}Sun Y, Wang S, Li Y et al., ``Ernie: Enhanced representation through knowledge integration,'' 2019, https://arxiv.org/abs/1904.09223.


\bibitem{ref32}Etzioni O, Cafarella M, Downey D et al., ``Unsupervised named-entity extraction from the web: An experimental study,'' Artificial intelligence., pp. 91-134., 2005.

\bibitem{ref33}Collins M, and Singer Y, ``Unsupervised models for named entity classification,'' 1999 Joint SIGDAT Conference on Empirical Methods in Natural Language Processing and Very Large Corpora., 1999.

\bibitem{ref34}Zhou G D, and Su J, ``Named entity recognition using an HMM-based chunk tagger,'' in Proceedings of the 40th Annual Meeting of the Association for Computational Linguistics., pp. 473-480, 2002.

\bibitem{ref35}Malouf R, ``Markov models for language-independent named entity recognition,'' COLING-02: The 6th Conference on Natural Language Learning 2002 (CoNLL-2002)., 2002.

\bibitem{ref36}Dai Z, Wang X, Ni P et al., ``Named Entity Recognition Using BERT BiLSTM CRF for Chinese Electronic Health Records,'' in 2019 12th international congress on image and signal processing, biomedical engineering and informatics (cisp-bmei). IEEE, pp. 1-5, 2019.


\bibitem{ref37}Liu W, Fu X, Zhang Y et al., ``Lexicon Enhanced Chinese Sequence Labelling Using BERT Adapter,'' 2021, https://arxiv.org/abs/2105.07148.


\bibitem{ref38}Chen Z Y, Chang C H, Chen Y P et al., ``UHop: An
unrestricted-hop relation extraction framework for
knowledge-based question answering,'' in Proceedings of NAACL-HLT., pp. 345-356, 2019.

\bibitem{ref39}Lan Y, Wang S, and Jiang J, ``Multi-hop knowledge base question answering with an iterative sequence matching model,'' in IEEE International Conference on Data Mining (ICDM)., pp. 359-368, 2019.

\bibitem{ref40}Chada R, and Natarajan P, ``FewshotQA: A simple framework for few-shot learning of question answering tasks using pre-trained text-to-text models,'' 2021, https://arxiv.org/abs/2109.01951.

\bibitem{ref41}
Hua Y, Li Y F, Haffari G et al., ``Few-shot Complex Knowledge Base Question Answering via Meta Reinforcement Learning,'' in Proceedings of the 2020 Conference on Empirical Methods in Natural Language Processing (EMNLP)., pp. 5827-5837, 2002.

\bibitem{ref42}
Wang X L, Li S C, Yang Z H et al., ``A Chinese KBQA System Based on Pre-trained Language Model,'' Journal of Shanxi University(Natural Science Edition)., vol. 43, pp. 955-962, 2020.(in Chinese)

\bibitem{ref43}Bast H, and Haussmann E, ``More accurate question answering on freebase,'' in Proceedings of the 24th ACM International on Conference on Information and Knowledge Management., pp. 1431-1440, 2015.

\bibitem{ref44}Unger C, and Cimiano P. Pythia, ``Natural language question answering over RDF: a graph data driven approach,'' in Proceedings of the 2014 ACM SIGMOD international conference on Management of data, pp. 313-324, 2014.

\bibitem{ref45}Talmor, Alon, and Jonathan Berant, ``The web as a knowledge-base for answering complex questions,'' 2018, https://arxiv.org/abs/1803.06643.

\bibitem{ref46}Xu K, Lai Y, Feng Y et al., ``Enhancing key-value memory neural networks for knowledge based question answering,'' in Proceedings of the 2019 Conference of the North American Chapter of the Association for Computational Linguistics: Human Language Technologies, vol. 1, pp. 2937-2947, 2019.

\end{thebibliography}
\end{document}